\documentclass{article}

\usepackage{wrapfig} 
\usepackage{booktabs} 
\usepackage{amsmath}
\usepackage{amsfonts}
\usepackage{comment}
\usepackage{footmisc}
\usepackage{mathrsfs}
\usepackage{multirow}
\usepackage{multicol}  
\usepackage{enumitem}
\usepackage{array}
\usepackage{float}
\usepackage{amssymb}
\usepackage{mathtools}
\usepackage{amsthm}
\usepackage{microtype}
\usepackage{graphicx}
\usepackage{subcaption}
\usepackage{bm}
\usepackage{dsfont}
\usepackage{adjustbox}
\usepackage{tcolorbox}
\usepackage{makecell}
\usepackage{hyperref}
\usepackage{url}
\usepackage{dsfont}
\usepackage{listings} 
\usepackage{makecell} 
\usepackage{caption}
\usepackage{wrapfig}
\usepackage{verbatim}

\usepackage{amsmath,amsfonts,bm}

\def\secref#1{\S\ref{#1}}

\def\eqref#1{equation~\ref{#1}}

\def\1{\bm{1}}

\DeclareMathAlphabet{\mathsfit}{\encodingdefault}{\sfdefault}{m}{sl}
\SetMathAlphabet{\mathsfit}{bold}{\encodingdefault}{\sfdefault}{bx}{n}

\def\eg{\emph{e.g.}} 

\def\ie{\emph{i.e.}}

\usepackage[capitalize,noabbrev]{cleveref}
\crefname{section}{\S}{\S\S}
\Crefname{section}{\S}{\S\S}
\crefname{subsection}{\S}{\S\S}
\Crefname{subsection}{\S}{\S\S}
\crefname{subsubsection}{\S}{\S\S}
\Crefname{subsubsection}{\S}{\S\S}
\usepackage[dvipsnames]{xcolor}

\newcommand{\ours}{\texttt{NV-CoT}}

\usepackage{color, colortbl}

\definecolor{lightCyan}{rgb}{0.925,1,1}

\definecolor{codegreen}{rgb}{0,0.6,0}
\definecolor{codegray}{rgb}{0.5,0.5,0.5}
\definecolor{codepurple}{rgb}{0.58,0,0.82}
\definecolor{backcolour}{rgb}{0.95,0.95,0.92}
\lstdefinestyle{mystyle}{
    backgroundcolor=\color{backcolour},   
    commentstyle=\color{codegreen},
    keywordstyle=\color{magenta},
    numberstyle=\tiny\color{codegray},
    stringstyle=\color{codepurple},
    basicstyle=\ttfamily\footnotesize,
    breakatwhitespace=false,         
    breaklines=true,                 
    captionpos=b,                    
    keepspaces=true,                 
    numbers=left,                    
    numbersep=5pt,                  
    showspaces=false,                
    showstringspaces=false,
    showtabs=false,                  
    tabsize=2
}

\PassOptionsToPackage{numbers,sort&compress}{natbib}

 \usepackage[main, final]{neurips_2026}

\usepackage[utf8]{inputenc} 
\usepackage[T1]{fontenc}    
\usepackage{hyperref}       
\usepackage{url}            
\usepackage{booktabs}       
\usepackage{amsfonts}       
\usepackage{nicefrac}       
\usepackage{marvosym}      
\usepackage{microtype}      
\usepackage{xcolor}         

\title{Thinking with Images as Continuous Actions: Numerical Visual Chain-of-Thought}

\author{Kesen Zhao\textsuperscript{1}\quad
        Beier Zhu\textsuperscript{2,\Letter}\quad
        Junbao Zhou\textsuperscript{1}\quad
         Xingyu Zhu\textsuperscript{3}\quad
        Zhongqi Yue\textsuperscript{4}\quad
        Hanwang Zhang\textsuperscript{1} \\
  $^{1}$NTU  \quad
  $^{2}$USTC \quad
  $^{3}$NUS \quad
  $^{4}$Microsoft Research\\
   Correspondence to: Beier Zhu \texttt{<beier.zhu@ustc.edu.cn>}\\
  }

\begin{document}

\maketitle

\begin{abstract}
Recent MLLMs increasingly rely on \emph{visual chain-of-thought} to perform region-grounded reasoning over images. However, existing approaches ground regions via either textified coordinates—causing modality mismatch and semantic fragmentation—or fixed-granularity patches that both limit precise region selection and often require non-trivial architectural changes. 
In this paper, we propose \texttt{N}umerical \texttt{V}isual \texttt{C}hain-of-\texttt{T}hought (\ours), a framework that enables MLLMs to reason over images using \emph{continuous} numerical coordinates. \ours~expands the MLLM action space from discrete vocabulary tokens to a continuous Euclidean space, allowing models to directly generate bounding-box coordinates as actions with only minimal architectural modification. The framework supports both supervised fine-tuning and reinforcement learning. In particular, we replace categorical token policies with a Gaussian (or Laplace) policy over coordinates and introduce stochasticity via reparameterized sampling, making \ours~fully compatible with GRPO-style policy optimization.
Extensive experiments on three benchmarks against 11 representative visual reasoning baselines demonstrate that \ours~significantly improves localization precision and final answer accuracy, while also accelerating training convergence, validating the effectiveness of continuous-action visual reasoning in MLLMs.
\textbf{Project page:} \href{https://kesenzhao.github.io/NV-CoT/}{\textcolor[RGB]{54,125,189}{kesenzhao.github.io/NV-CoT/}}
\end{abstract}

\begin{figure*}[h]
    \centering
    \includegraphics[width=1\linewidth]{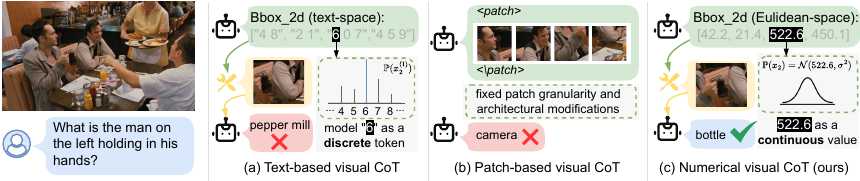}
    \caption{
Comparison of different paradigms for thinking with images.
(a) Text-based approaches represent localized regions as discrete coordinate tokens, leading to modality mismatch and fragmented semantics.
(b) Patch-based approaches reason directly over fine-grained visual tokens but are constrained by the fixed spatial granularity of the vision backbone.
(c) Our \ours~predicts region coordinates in continuous space, enabling flexible and precise localization.}
    \label{fig:overview}
\end{figure*}

\section{Introduction}
\label{sec:intro}

Recent advances in multimodal large language models (MLLMs) have enabled deeper reasoning over visual inputs via ``thinking with images'' (\ie, visual chain-of-thought)~\cite{zheng2025deepeyes, shao2024visual, zhao2025unsupervised, li2025imagine, su2025openthinkimg}. In this paradigm, models first localize task-relevant image regions aligned with the textual query. They then conduct multi-step reasoning by explicitly grounding intermediate inferences on these detected regions. Such region-grounded visual reasoning is a fundamental capability for MLLMs and underpins a broad range of downstream applications.

To align with the text-based output interface, many existing MLLMs~\cite{liu2025visionreasoner, bai2025qwen2, zhu2025internvl3} serialize localized regions as bounding-box coordinates \textbf{in text}, \eg, $[\text{``}x_1\text{''},\text{``}y_1\text{''},\text{``}x_2\text{''},\text{``}y_2\text{''}]$, and then invoke a cropping tool to extract the corresponding patch (\cref{fig:overview} (a)).
Despite its simplicity, this design faces two fundamental issues. (1) Modality mismatch: coordinates are continuous in the visual world but are predicted as discrete text tokens and typically trained with cross-entropy, which ignores geometric proximity (\eg, predicting $x_1{=}\text{``}3.2\text{''}$ may be penalized similarly to $x_1{=}\text{``}4.1\text{''}$ when the target is $x_1^\ast{=}\text{``}3.1\text{''}$, since both predictions differ from the target by only a single discrete token under token-level cross-entropy). (2) Semantic fragmentation: numbers are tokenized into multiple unrelated sub-tokens, making numerical comparison and reasoning brittle and prone to hallucination (\eg, confusing whether $\text{``}3.11\text{''}$ or $\text{``}3.9\text{''}$ is larger).

\begin{figure*}[t]
  \centering
  \begin{minipage}[c]{0.5\textwidth}
    \includegraphics[width=\linewidth]{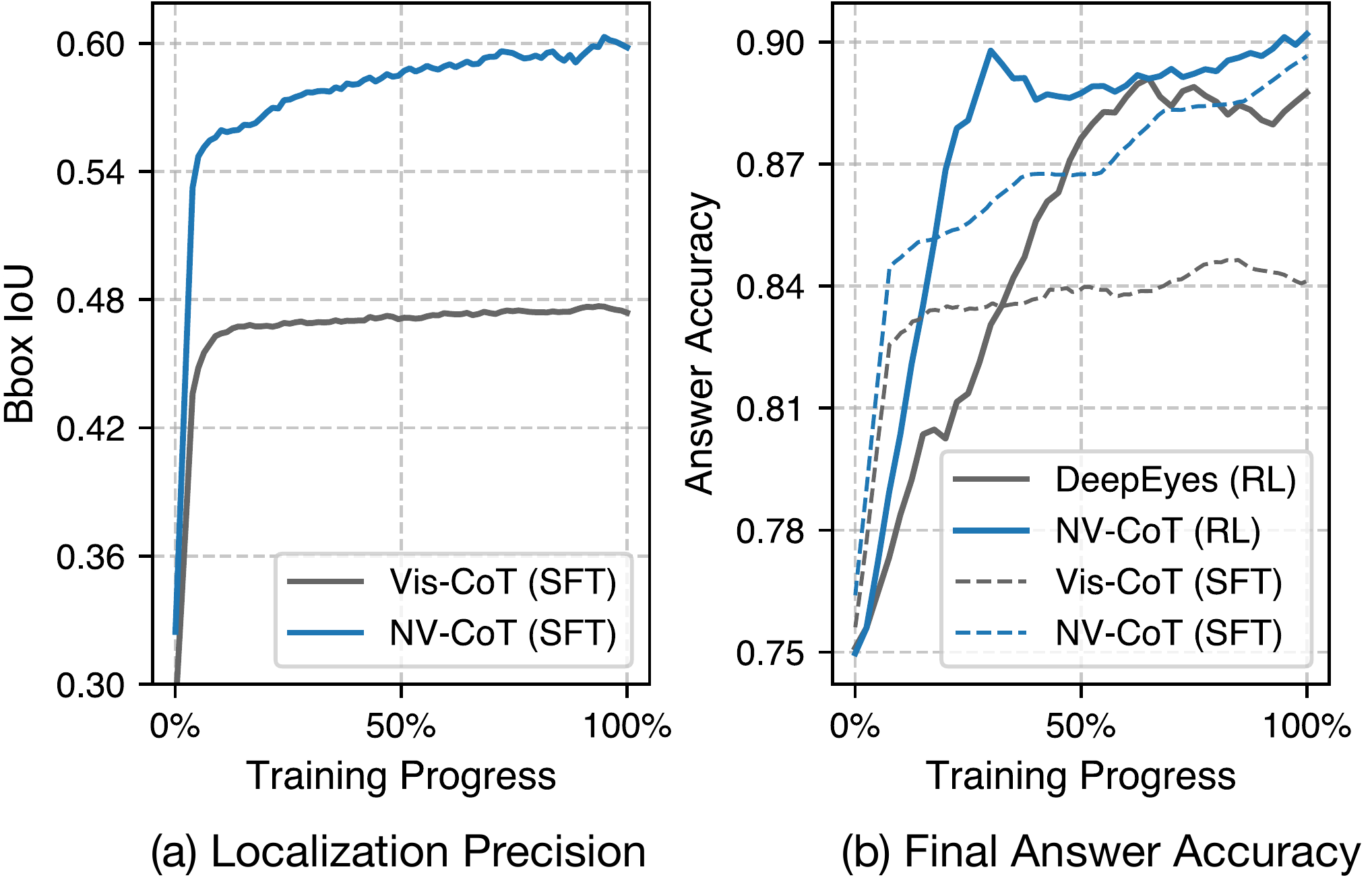}
  \end{minipage}
  \hfill
  \begin{minipage}[c]{0.45\textwidth}
    \small
    \caption{Our \ours~outperforms text-based visual CoT models (Vis-CoT~\cite{shao2024visual} and DeepEyes~\cite{zheng2025deepeyes}) in \textbf{localization precision}, \textbf{answer accuracy}, and \textbf{convergence speed} across both SFT and RL. SFT-based models are evaluated on the Vis-CoT-363K dataset~\cite{shao2024visual}, where ground-truth bounding boxes are available, while RL-based models are evaluated on the DeepEyes-47K dataset~\cite{zheng2025deepeyes}. We only replace the text-space discrete coordinate objective with our Euclidean-space continuous one, while keeping all other training configurations unchanged for a fair comparison.}
    \label{fig:efficiency}
  \end{minipage}
\end{figure*}

Several recent works attempt to alleviate these issues by enabling MLLMs to operate directly on fine-grained visual patches (\cref{fig:overview} (b)). For example, LVR \cite{li2025latent} performs reasoning in the visual embedding space, while PaDT~\cite{su2025patch} generates indices corresponding to original visual patch embeddings.
Nevertheless, these approaches are fundamentally constrained by the fixed granularity of the vision backbone, as the predefined patch partitioning limits flexible and precise region selection. In addition, patch-generation or patch-indexing mechanisms often require non-trivial architectural changes, which weakens modularity and limits general applicability across MLLMs.

To avoid brittle textified coordinates with only minimal architectural changes, we propose \texttt{N}umerical  \texttt{V}isual \texttt{C}hain-\texttt{o}f-\texttt{T}hought (\ours), which expands the MLLM action space from discrete vocabulary tokens to a continuous Euclidean space, enabling the model to directly generate numerical bounding box coordinates as actions (\cref{fig:overview} (c)). 
Specifically, we extend the standard LLM head with four coordinate outputs that directly predict the numerical bounding-box coordinates (\eg, $[x_1, y_1, x_2, y_2]$\footnote{We write text-based coordinates as $[\text{``}x_1\text{''},\text{``}y_1\text{''},\text{``}x_2\text{''},\text{``}y_2\text{''}]$ and numerical coordinates as $[x_1,y_1,x_2,y_2]$.}). 
Notably, \ours~is applicable to both supervised fine-tuning (SFT) and reinforcement learning (RL) stages.
In the SFT stage, we replace token-level cross-entropy with a regression loss to supervise continuous bounding-box coordinates. However, extending this idea to the RL stage is non-trivial. 
Most LLM-oriented policy optimization methods parameterize the policy as a token-level categorical distribution over the vocabulary, which naturally supports discrete actions but does not directly accommodate continuous coordinate outputs.
{Furthermore,  deterministic coordinate prediction lacks the stochasticity required by GRPO-style algorithms for advantage estimation and policy exploration.}
To address these challenges, we replace the categorical policy with a Gaussian policy that explicitly models continuous coordinate regression. Stochasticity is introduced via the reparameterization trick, where the model predicts both the mean and standard deviation of the bounding box distribution and samples coordinates accordingly. This design enables efficient exploration and stable gradient propagation, while also providing closed-form likelihoods for computing importance ratios and KL regularization, making \ours~compatible with GRPO-style RL frameworks. Finally, we also provide a Laplace-policy variant of \ours, motivated by the empirical robustness and favorable performance of $\ell_1$-type losses for localization; it serves as a drop-in replacement for the Gaussian policy under the same GRPO pipeline.

In \cref{fig:efficiency}, we plot bounding-box precision (IoU under the SFT setting with ground-truth coordinates) and final answer accuracy over the course of training. The results show that \ours~consistently improves \textbf{localization accuracy}, \textbf{final answer accuracy}, and \textbf{convergence speed}.
Extensive experiments on $V^*$ Bench~\cite{wu2024v}, HR-Bench 4K~\cite{wang2025divide}, and HR-Bench 8K~\cite{wang2025divide} demonstrate that \ours~consistently improves visual reasoning performance. For example, on $V^*$ Bench, \ours\texttt{-7B} outperforms LVR-7B (patch-based visual CoT) and DeepEyes-7B (text-based visual CoT) by $7.8\%$ and $2.6\%$, respectively (\cref{table:overall}). Beyond high-resolution benchmarks, \ours~also outperforms baselines on general multimodal benchmarks, including MMMU-Pro~\cite{yue2024mmmu_pro}, ChartQA~\cite{masry2022chartqa}, and OCRBenchv2~\cite{fu2025ocrbenchv2} (\cref{tab:mmbench}). It also retains performance on pure-text reasoning tasks (\cref{tab:pure_text}). On grounding and counting benchmarks, \ours~outperforms both Qwen2.5-VL-7B and DeepEyes-7B across the RefCOCO series, COCO detection, CountBenchQA, and TallyQA (\cref{tab:grounding_counting}).

\section{Related Work}
\label{sec:relatedwork}

\textbf{Multimodal large language models.} By integrating language modeling capabilities with visual understanding, MLLMs enable complex vision-language tasks.
Representative early models such as BLIP-2~\cite{li2023blip} and LLaVA~\cite{liu2023visual, liu2024improved} align vision and language by projecting image into the latent space of a frozen LLM via a query transformer or a lightweight projector.
Subsequent strong open-source models, including the LLaVA family~\cite{liu2024llava, guo2024llava, zhang2025llava, lin2024video, li2023llava, li2024llava}, Qwen-VL~\cite{bai2023qwen, wang2024qwen2, bai2025qwen2}, and InternVL~\cite{chen2024internvl, lu2025internvl}, further improve visual resolution handling and scalability. However, despite their strong perception and alignment capabilities, these models generally lack explicit reasoning mechanisms.

\textbf{Thinking with Images (visual CoT).} 
Incorporating CoT reasoning to enhance visual understanding has been actively explored in recent MLLMs~\cite{wang2025pixel, zhang2025chain}.
Early text-based approaches~\cite{liu2025visionreasoner, bai2025qwen2, zhu2025internvl3} serialize localized image regions as bounding-box coordinates in text, and subsequently invoke cropping tools to extract the corresponding visual patches. While effective, this paradigm suffers from modality mismatch and semantic fragmentation.
More recent patch-based methods~\cite{li2025latent, su2025patch, huang2025visualtoolagent} instead operate directly on fine-grained visual patches, avoiding explicit textual serialization. However, these approaches are fundamentally constrained by the fixed spatial granularity of the vision backbone, limiting their flexibility in region selection and reasoning precision.
In contrast, \ours~expands the MLLM action space from discrete vocabulary tokens to a continuous Euclidean space, enabling the model to directly predict numerical bounding-box coordinates as continuous actions. This design naturally bridges visual perception and reasoning, without relying on fixed patch partitions. 

To incorporate CoT to MLLMs, existing methods adopt either supervised fine-tuning~\cite{zhang2025cmmcot, wang2025vgr, chung2025don, shao2024visual} or reinforcement learning~\cite{zhang2025chain, su2025openthinkimg, zheng2025deepeyes, zhao2025unsupervised}.
Our approach is compatible with both paradigms, providing a unified framework for 
learning continuous visual actions.
\section{Method}
\label{sec:method}

Given a query $q$ and an image $I_0$, a visual CoT MLLM $\pi_\theta$ performs region-grounded reasoning by  selecting a region of interest and invoking a zoom-in function for fine-grained inspection. Existing visual CoT methods localize regions by generating \textbf{text-form} box coordinates and then cropping the image for feedback. We instead cast localization as a continuous-action problem in Euclidean space: the MLLM directly outputs \textbf{numerical coordinates}.
Our method proceeds in three steps. 
First, we expand the MLLM vocabulary with a continuous coordinate action space (\secref{sec:actionspace}), enabling numerical box generation.
Second, we instantiate \ours~for SFT with coordinate regression when box supervision is available (\secref{sec:sft}).
Third, we extend \ours~to GRPO by introducing a stochastic Gaussian policy with closed-form importance ratios and KL regularization (\secref{sec:rl}).
Finally, we provide a Laplace-policy variant as a drop-in replacement, motivated by the empirical preference of $\ell_1$-type losses for localization (\secref{sec:laplace}).

\subsection{Expanded Actions for Continuous Coordinates}\label{sec:actionspace}
To enable numerical coordinate prediction, we augment the standard LLM output head over the vocabulary $\mathcal{V}$ with a continuous action space $\mathcal{A}=\mathbb{R}^4\times\mathbb{R}_+$: four coordinate heads parameterized by
$W=[\mathbf{w}_{x_1},\mathbf{w}_{y_1},\mathbf{w}_{x_2},\mathbf{w}_{y_2}]^\top \in \mathbb{R}^{4\times D}$ to predict the Gaussian mean, and a fifth head $\mathbf{w}_\sigma \in \mathbb{R}^{D}$ to predict a shared standard deviation, yielding a Gaussian policy for continuous localization.
Let $\mathbf{h}\in\mathbb{R}^{D}$ denote the final hidden representation before the coordinate decoding step.
We obtain the mean and shared standard deviation as
\begin{equation}\label{eq:std}
    \bm{\mu} = [x_1,y_1,x_2,y_2]^\top = W\mathbf{h}, \qquad \sigma = \mathrm{max}(\mathbf{w}_\sigma^\top \mathbf{h}, \varepsilon),
\end{equation}
where $\varepsilon > 0$ is a small positive offset to ensure $\sigma>0$.
This forms a Gaussian policy $\mathcal{N}(\bm{\mu}, \sigma^2 I)$. 
Accordingly, we augment the parameter set as $\theta \leftarrow \theta \cup \{W, \mathbf{w}_{\sigma}\}$. 

The introduction of $\sigma$ is crucial: while deterministic coordinate suffices for SFT, GRPO-style RL requires stochastic policies to enable sampling and exploration.  
Note that our modification to the MLLM is minimal, as we only introduce five lightweight linear heads, in contrast to patch-based visual CoT methods that require more architectural changes.

\subsection{\ours~for Supervised Fine-Tuning}\label{sec:sft}
We begin by applying our \ours~to supervised fine-tuning, where intermediate bounding-box coordinates in the reasoning process are available~\cite{chung2025don, shao2024visual}.
Let $\mathbf{b^\ast}=[x_1^\ast,y_1^\ast,x_2^\ast,y_2^\ast]^\top$ denote the ground-truth bounding box for a zoom-in step.
During SFT, we supervise the coordinate outputs in $\mathcal{A}$ using an $\ell_2^2$ regression loss, while optimizing the discrete token outputs in the original LLM head $\mathcal{V}$ with the standard cross-entropy loss.
\begin{equation}
\label{eq:sft}
\mathcal{L}_{\mathsf{SFT}}(\theta)
=
\sum_{t\in \mathcal{T}_{\mathcal{V}}}\!\!\Big(-\log \pi_\theta(o_t \mid o_{<t}, q, I_0)\Big) \notag \;+\;\lambda 
\sum_{t\in \mathcal{T}_{\mathcal{A}}}\!\! \|\bm{\mu}_t-\mathbf{b}_t^\ast\|_2^2,
\end{equation}
where $\mathcal{T}_{\mathcal{V}}=\{t:o_t\in\mathcal{V}\}$ and $\mathcal{T}_{\mathcal{A}}=\{t:o_t\in\mathcal{A}\}$ denote textual token and coordinate-action positions, respectively; $o_t$ is the output of the LLM, and $\lambda>0$ is a hyperparameter; $\sigma$ is not supervised in SFT and is only used for RL.

\subsection{\ours~for Reinforcement Learning}\label{sec:rl}
Bounding-box annotations for SFT are costly and often unavailable at scale.
As a result, visual CoT models are commonly trained with RL (GRPO~\citet{shao2024deepseekmathpushinglimitsmathematical}) using task-level feedback (\eg, final answer) as the reward signal, without requiring box-level supervision.
Extending \ours~from SFT to this RL setting, however, is non-trivial: RL needs stochastic sampling and well-defined importance ratios, whereas standard LLM policies are token-level categorical distributions that do not accommodate continuous coordinate actions.
To this end, we formulate continuous region localization as a Gaussian policy over the action space $\mathcal{A}$. 

\textbf{Rollouts.} GRPO samples a group of $G$ trajectories per prompt. With our Gaussian policy, we independently sample coordinate actions as
$
\mathbf{b}^{(i)} \sim \mathcal{N}(\bm{\mu}, \sigma^2 I), \; i=1,\ldots,G,
$
To enable low-variance gradient estimation, we implement this sampling using the reparameterization trick:
\begin{equation}\label{eq:reparam}
\mathbf{b}^{(i)} = \bm{\mu} + \sigma \bm{\epsilon}^{(i)}, \quad \bm{\epsilon}^{(i)} \sim \mathcal{N}(\mathbf{0}, I).
\end{equation}
We then use $\mathbf{b}^{(i)}$ to crop into the image for the $i$-th trajectory.

\textbf{Importance ratio.}
GRPO optimizes the policy via an importance ratio between the current policy $\pi_\theta$ and a reference policy $\pi_{\theta_{\mathsf{old}}}$.
For each output $o^{(i)}_t$, the ratio is defined as
\begin{equation}
r^{(i)}_{t}(\theta)
=
\frac{\pi_\theta(o^{(i)}_{t}\mid q, I_0, o^{(i)}_{<t})}
     {\pi_{\theta_{\mathsf{old}}}(o^{(i)}_t\mid q,I_0, o^{(i)}_{<t})}.
\end{equation}
When $o^{(i)}_{t}\in\mathcal{V}$ is a text token, $r^{(i)}_{t}(\theta)$ reduces to the standard likelihood ratio of categorical policies used in  GRPO.
When $o^{(i)}_{t} \in\mathcal{A}$ corresponds to a continuous coordinate action $\mathbf{b}^{(i)}_{t}$, we compute the ratio using Gaussian likelihoods:
\begin{equation}
r^{(i)}_{t}(\theta)
=
\frac{\mathcal{N}(\mathbf{b}^{(i)}_{t}\mid \bm{\mu}_\theta,\sigma_\theta^2 I)}
     {\mathcal{N}(\mathbf{b}^{(i)}_{t}\mid \bm{\mu}_{\theta_{\mathsf{old}}},\sigma_{\theta_{\mathsf{old}}}^2 I)}, \ \text{where} \
\mathcal{N}(\mathbf{b}\mid \bm{\mu},\sigma^2 I)
=
\frac{1}{(2\pi\sigma^2)^{d/2}}
\exp\!\left(-\frac{\|\mathbf{b}-\bm{\mu}\|_2^2}{2\sigma^2}\right),
\end{equation}
Setting $d=4$ and substituting this density into the ratio gives the importance ratio in terms of squared Mahalanobis distances:
\begin{equation}\label{eq:ratio}
r^{(i)}_{t}(\theta)
=\frac{\sigma^4_{\theta_{\mathsf{old}}}}{\sigma^4_{\theta}}
\exp\!\left(
-\frac{\|\mathbf{b}^{(i)}_{t}-\bm{\mu}_\theta\|_2^2}{2\sigma_\theta^2}
+
\frac{\|\mathbf{b}^{(i)}_{t}-\bm{\mu}_{\theta_{\text{old}}}\|_2^2}{2\sigma_{\theta_{\text{old}}}^2}
\right).
\end{equation}

\noindent\textbf{KL penalty.} Thanks to the closed-form KL divergence of Gaussian distributions, this penalty is analytically tractable for continuous actions $\mathcal{A}$:
\begin{equation}
\mathrm{KL}\!\left(\pi_\theta \,\|\, \pi_{\mathsf{ref}}\right)
=
\tfrac{\|\bm{\mu}_\theta-\bm{\mu}_{\mathsf{ref}}\|_2^2}{2\sigma_{\mathsf{ref}}^{2}}
+
2\!\left(
\tfrac{\sigma_\theta^2}{\sigma_{\mathsf{ref}}^2}
-\log \sigma_\theta^2
\right)
+ C,
\end{equation}
where $C$ is a constant. However, $\sigma_{\mathsf{ref}}$ is unknown since $\sigma$ is not supervised in SFT.
Therefore, we drop the variance-related term and  only constrain the mean: $\mathrm{KL}_{\bm{\mu}}\!\left(\pi_\theta \,\|\, \pi_{\mathsf{ref}}\right)
=\|\bm{\mu}_\theta-\bm{\mu}_{\mathsf{ref}}\|_2^2$. If SFT stage is unavailable (and thus no reference mean $\bm{\mu}_{\mathsf{ref}}$ can be obtained), we omit this KL regularization term.
For textual token outputs in $\mathcal{V}$, we retain the standard token-level KL divergence.

\noindent\textbf{Reward.} Following DeepEyes~\cite{zheng2025deepeyes}, we adopt an outcome-driven reward that combines answer correctness, format validity, and a conditional zoom-in use bonus.  
Formally, for a trajectory $o^{(i)}$, the total reward is
\begin{equation}\label{eq:award}
R(o^{(i)}) \;=\; R_{\mathsf{acc}}(o^{(i)}) \;+\; R_{\mathsf{fmt}}(o^{(i)}) \;+\;   R_{\mathsf{zoom}}(o^{(i)}),
\end{equation}
where the zoom-in bonus is activated only when the trajectory is correct and invokes the visual grounding  action at least once, encouraging region of interest identification.  

\noindent\textbf{Optimization.} Finally, given a question $q$ and an input image $I_0$, we sample a group of $G$ trajectories $\{o^{(1)}, \ldots, o^{(G)}\}$ from $\pi_{\theta_{\mathsf{old}}}$, each associated with intermediate bounding boxes $\{\mathbf{b}^{(1)}, \ldots, \mathbf{b}^{(G)}\}$ (via Eq.~(\ref{eq:reparam})). We compute trajectory-level rewards $\mathbf{R}=\{R(o^{(1)}),\dots, R(o^{(G)})\}$ using Eq.~(\ref{eq:award}) and use the normalized advantage for all steps in the $i$-th trajectory:
$A_t^{(i)}=\frac{R(o^{(i)})-\mathsf{mean}(\mathbf{R})}{\mathsf{std}(\mathbf{R})}$. Given the importance ratio $r_t^{(i)}(\theta)$ and the KL term $\mathrm{KL}(\pi_\theta \,\|\, \pi_{\mathrm{ref}})$ defined above, we optimize the objective:
\begin{align} 
\mathcal{L}_{\mathsf{GRPO}}(\theta)  &=\mathbb{E}_{q,I_0,\left\{o^{(i)}\right\}_{i=1}^G \sim \pi_{\theta_{\mathsf{old}}} } \notag \Big[\frac{1}{GT} \sum_{i=1}^G\sum_{t=1}^{T} \min \left(r_{t}^{(i)}(\theta) {A}_{t}^{(i)}, \mathrm{clip}\big(r_{t}^{(i)}(\theta), 1-\varepsilon, 1+\varepsilon\big) {A}_{t}^{(i)}\right) \notag \\ 
& - \beta \mathrm{KL}(\pi_\theta \| \pi_{\mathsf{ref}})\Big],\quad \text{where } \varepsilon>0 \text{ is the clipping range}.
\end{align}

\noindent\textbf{Inference.} During inference, we disable stochasticity, crop using the coordinate mean $\bm{\mu}$, and express all iterative zoom-in boxes in the original image's coordinate frame.

\subsection{Extending \ours~with Laplace Policy}\label{sec:laplace}

In prediction tasks such as keypoint localization and bounding-box regression, it is well known that $\ell_1$ loss often outperforms $\ell_2^2$ loss due to its robustness to outliers and sharper error profiles~\cite{Girshick_2015_ICCV,sun2018integral}.
From a probabilistic perspective, this corresponds to modeling prediction errors with a Laplace distribution rather than a Gaussian distribution.
Motivated by this observation, we further extend \ours~by replacing the Gaussian policy with a Laplace policy for continuous localization.

\paragraph{Laplace policy.}
Specifically, we model the coordinate action as $\mathbf{b} \sim \mathrm{Laplace}(\bm{\mu}, \alpha)$, whose probability density function is
\begin{equation}
p(\mathbf{b}\mid \bm{\mu},  \alpha)
=
\frac{1}{(2\alpha)^4}
\exp\!\left(
-\frac{\|\mathbf{b}-\bm{\mu}\|_1}{ \alpha}
\right),
\end{equation}
where $\bm{\mu}\in\mathbb{R}^4$ denotes the predicted coordinate mean and $\alpha>0$ is a shared scale parameter.
Equivalently, this corresponds to assuming independent dimensions, under which the Laplace likelihood factorizes across coordinates.

\paragraph{SFT objective.}
Under the Laplace assumption, maximizing the log-likelihood of ground-truth coordinates is equivalent to minimizing an $\ell_1$ loss, since
$-\log p(\mathbf{b}^\ast\mid \bm{\mu},  \alpha)=\tfrac{1}{ \alpha}\|\bm{\mu}-\mathbf{b}^\ast\|_1 + \mathrm{const}$.
Thus, in the SFT stage, we replace the $\ell_2^2$ loss Eq.~(\ref{eq:sft}) with
\begin{equation}
\mathcal{L}_{\mathsf{SFT}}^{\mathsf{Lap}}(\theta)
=
\sum_{t\in \mathcal{Y}_\mathcal{A}}\|\bm{\mu}_t-\mathbf{b}^\ast_t\|_1,
\end{equation}
which encourages sharper and more robust localization.

\paragraph{RL with Laplace policy.}
Analogous to the Gaussian case, we parameterize the Laplace policy with a location parameter $\bm{\mu}$ and a scale parameter $\alpha$, and employ reparameterized sampling to enable low-variance gradient estimation.
Specifically, a coordinate action is sampled as
\begin{equation}
\mathbf{b}
=
\bm{\mu}
+
 \alpha\,\mathbf{s}\odot\boldsymbol{\epsilon},
\quad
\mathbf{s}\sim\mathrm{Rademacher}(\pm1),\;
\boldsymbol{\epsilon}\sim\mathrm{Exp}(\mathbf{1}),
\end{equation}
where $\odot$ denotes element-wise multiplication. Here, $\mathbf{s}$ is a random sign vector with i.i.d.\ entries in $\{\pm 1\}$, and $\boldsymbol{\epsilon}$ has i.i.d.\ unit-rate exponential entries.
For a sampled action $\mathbf{b}_{t}^{(i)}$, the corresponding importance ratio used in GRPO is
\begin{equation}
r_{t}^{(i)}(\theta)
=
 \frac{ \alpha^{4}_{\theta_{\mathsf{old}}}}{ \alpha_\theta^{4}}
\exp\!\left(
-\frac{\|\mathbf{b}_{t}^{(i)}-\bm{\mu}_\theta\|_1}{ \alpha_\theta}
+
\frac{\|\mathbf{b}_{t}^{(i)}-\bm{\mu}_{\theta_{\mathsf{old}}}\|_1}{ \alpha_{\theta_{\mathsf{old}}}}
\right),
\end{equation}
which can be computed analytically. The closed-form KL penalty can be derived analogously, and the overall GRPO procedure follows the same pipeline as in the Gaussian-policy case.

\section{Experiments}
\label{sec:experiments}
In \secref{sec:setup}, we describe the experimental setup. In \secref{sec:main}, we report the main results on high-resolution benchmarks. In \secref{sec:general_results}, we present results beyond high-resolution benchmarks, including general multimodal, pure-text, grounding, and counting tasks. In \secref{sec:ablation}, we present ablation studies and analyses of uncertainty, the training coefficient, training dynamics, and qualitative grounding.
\subsection{Setup}
\label{sec:setup}


\textbf{Baselines.} 
We compare against 11 state-of-the-art MLLM baselines, grouped into four categories. (1) Open-source MLLMs: Qwen2.5-VL-\{7B, 32B\}~\cite{bai2025qwen2} and LLaVA-OneVision~\cite{li2024llava}. (2) Training-free visual CoT: DyFo-7B~\cite{li2025dyfo} (3) SFT-based visual CoT: Vis-CoT-7B~\cite{shao2024visual}, ZoomEye-7B~\cite{shen2025zoomeye} and RAP-7B~\cite{wang2025retrieval}. (4) RL-based visual CoT: UV-CoT-7B~\cite{zhao2025unsupervised}, DeepEyes-7B~\cite{zheng2025deepeyes}, PaDT~\cite{su2025patch} and LVR~\cite{li2025latent}.
Our \ours~is applicable to both SFT and RL settings. In this work, we build the SFT-based model upon Vis-CoT~\cite{shao2024visual} and the RL-based model upon DeepEyes~\cite{zheng2025deepeyes}

\begin{table*}[t] 
\center
\fontsize {9}{10} \selectfont
\setlength{\tabcolsep}{6pt}
\caption{Results on high-resolution benchmarks. \textbf{Best} results are in bold. Our \ours~consistently improves performance under both SFT and RL settings. We use \ours~(SFT) with an $\ell_1$ loss and \ours~(RL) with a Laplace policy.
\textbf{DeepEyes-7B$^\dag$} and \textbf{\ours$^\dag$} denote results under multi-step tool use (multiple zoom-in calls). For a fair comparison, all other visual CoT methods are evaluated with a single zoom-in call, since some baselines do not support multiple tool invocations.
}
\label{table:overall}
\begin{tabular}{l|ccc|ccc|ccc}
\toprule[1pt]
\multirow{2}{*}{Model} & \multicolumn{3}{c|}{ $V^*$ Bench} & \multicolumn{3}{c|}{HR Bench 4K} & \multicolumn{3}{c}{HR Bench 8K} \\ \cmidrule{2-10}
& Attr & Spatial & Overall & FSP & FCP & Overall & FSP & FCP & Overall \\ \midrule
LLaVA-OneVision-7B  & 75.7 & 75.0 & 75.4 & 71.7 & 53.3 & 62.5 & 69.1 & 52.1 & 60.6 \\
Qwen2.5-VL-7B & 73.9 & 67.1 & 71.2 & 84.7 & 52.9 & 68.8 & 78.7 & 51.8 & 65.3 \\
Qwen2.5-VL-32B  & 87.8 & 88.1 & 87.9 & 89.4 & 58.3 & 73.9 & 83.6 & 56.9 & 70.3\\ \midrule
DyFo-7B & 80.0 & 82.9 & 81.2 & - & - & - & - & - & - \\
PaDT-7B & 75.7 & 73.6 & 74.9 & 86.0 & 54.8 & 70.4 & 79.0 & 53.8 & 66.4\\
LVR-7B & 84.4 & 76.3 & 81.2 & 88.3 & 56.8 & 72.5 & 82.5 & 54.5 & 68.5 \\ 
UV-CoT-7B & 82.6 & 76.3 & 80.1 & 79.8 & 54.5 & 67.1 & 76.8 & 54.3 & 65.5 \\  
ZoomEye-7B & 93.9 & 85.5 & 90.6 & 84.3 & 55.0 & 69.6 & 88.5 & 50.0 & 69.3 \\  
RAP-7B & 87.0 & 78.9 & 83.8 & 76.3 & 50.5 & 63.4 & 69.0 & 46.8 & 57.9 \\ \midrule
Vis-CoT-7B & 80.9 & 75.0 & 78.5 & 79.0 & 54.3 & 66.6 & 75.5 & 53.0 & 64.3 \\ 
\rowcolor{lightCyan} \quad + \ours~(SFT) & 84.3 & 78.9 & 82.2 & 82.0 & 55.3 & 68.6 & 77.3 & 54.5 & 65.9 \\ 
\quad $\Delta$   & +3.5 & +3.9 & +3.7 & +3.0 & +1.0 & +2.0 & +1.8 & +1.5 & +1.6 \\ \midrule
DeepEyes-7B & 87.8 & 84.2 & 86.4 & 89.3 & 57.8 & 73.5 & 84.3 & 56.5 & 70.4 \\
\rowcolor{lightCyan} \quad + \ours~(RL) & 90.4 & 86.8 & 89.0 & 91.0 & 58.5 & 74.8 & 85.8 & 58.3 & 72.0 \\ 
 \quad $\Delta$   & +2.6 & +2.6 & +2.6 & +1.7 & +0.7 & +1.3 & +1.5 & +1.8 & +1.6 \\ \midrule
 DeepEyes-7B$^\dag$ & 91.3 & 88.2 & 90.1 & 91.3 & 59.0 & 75.2 & 86.8 & 58.5 & 72.7 \\
\rowcolor{lightCyan} \quad + \ours$^\dag$~(RL) & \textbf{93.0} & \textbf{89.5} & \textbf{91.6} & \textbf{91.8} & \textbf{60.3} & \textbf{76.0} &\textbf{87.5} & \textbf{60.0} & \textbf{73.8} \\ 
 \quad $\Delta$   & +1.7 & +1.3 & +1.5 & +0.5 & +1.3 & +0.8 & +0.7 & +1.5 & +1.1 \\ \bottomrule[1pt]
\end{tabular}
\end{table*}

\textbf{Datasets.} For fair comparison, we use the \textit{same training data} as the corresponding backbones (\eg, DeepEyes-7B and Vis-CoT-7B). For evaluation, we consider four categories:
\textit{(1) High-resolution visual reasoning.} We evaluate on $V^*$ Bench~\cite{wu2024v}, which measures attribute (Attr) and spatial reasoning (Spatial), and HR-Bench 4K/8K~\cite{wang2025divide}, which assesses fine-grained spatial perception (FSP) and fine-grained comparison perception (FCP) across resolutions.
\textit{(2) General multimodal understanding.} We use MMMU-Pro~\cite{yue2024mmmu_pro} for multidisciplinary understanding, ChartQA~\cite{masry2022chartqa} for chart reasoning, and OCRBenchv2~\cite{fu2025ocrbenchv2} for visual text reasoning.
\textit{(3) Grounding and counting.} We evaluate referring-expression grounding on RefCOCO, RefCOCO+, and RefCOCOg~\cite{yu2016modeling,mao2016generation}, object detection on COCO~\cite{lin2014coco}, and visual counting on CountBenchQA~\cite{beyer2024paligemma} and TallyQA~\cite{acharya2019tallyqa}.
\textit{(4) Pure-text reasoning.} We evaluate language and mathematical reasoning on MMLU~\cite{hendrycks2021measuring} and GSM8K~\cite{cobbe2021training}, respectively.

\textbf{Evaluation protocols.} For a fair comparison under identical inference scaling, we evaluate all models with a single zoom-in per query, since some baselines (\eg, Vis-CoT~\cite{shao2024visual} and UV-CoT~\cite{zhao2025unsupervised}) do not support iterative tool invocations. For methods that do support multiple tool calls (\eg, DeepEyes~\cite{zheng2025deepeyes} and our \ours~(RL)), we additionally report results under multiple tool use.

\textbf{Implementation details.} To ensure a fair comparison, we \textit{keep the training configurations identical} to those of the backbone methods (DeepEyes-7B~\cite{zheng2025deepeyes} and Vis-CoT-7B~\cite{shao2024visual}), and only replace text-form coordinate prediction with our continuous action formulation. Specifically, for SFT, we fine-tune LLaVA for one epoch with a batch size of 128. 
We use the Adam optimizer with zero weight decay, a learning rate of $2\times10^{-5}$, and a cosine learning-rate scheduler. $\lambda$ in Eq.~(\ref{eq:sft}) is set to $0.3$. 
For RL, we train the RL-based model on top of Qwen2.5-VL-7B using GRPO for 80 iterations on 32 A100 GPUs. 
We set the batch size to 256 and the GRPO group size to $G=16$. 
The KL coefficient is set to $\beta=0$.

\subsection{Main Results on High-Resolution Benchmarks}
\label{sec:main}
The overall performance comparisons are reported in \cref{table:overall}, leading to the following observations:

\textbf{(1)} Our \ours~consistently outperforms the backbone models across all benchmarks under both SFT and RL.
Specifically, under SFT, \ours~improves over Vis-CoT-7B by $+3.7\%$ overall on $V^*$ Bench, $+2.0\%$ on HR-Bench 4K, and $+1.6\%$ on HR-Bench 8K.
Under RL, \ours~achieves further gains over DeepEyes-7B, with overall improvements of $+2.6\%$, $+1.3\%$, and $+1.6\%$ on the three benchmarks, respectively.
These results suggest that replacing text-based bounding-box prediction with our numerical one benefits both SFT and RL training paradigms.

\textbf{(2)} We observe a clear paradigm advantage of zoom-in visual CoT over patch-based reasoning.  
The backbone DeepEyes-7B and our \ours~(RL) consistently outperform patch-based baselines such as PaDT-7B and LVR-7B across all benchmarks, suggesting that region selection via zoom-in is more effective than reasoning over fixed-granularity patches.

\textbf{(3)} Methods that perform thinking with images substantially outperform basic MLLMs such as LLaVA-OneVision-7B and Qwen2.5-VL-7B.
Notably, despite being built on a 7B backbone, \ours~achieves performance that surpasses Qwen2.5-VL-32B across all benchmarks.
This highlights the importance of region-grounded visual reasoning in improving multimodal understanding.

\textbf{(4)} Allowing multiple zoom-in tool calls further improves ``thinking with images'' performance, indicating that iterative region selection and refinement is beneficial for fine-grained visual reasoning.
Notably, DeepEyes-7B$^\dag$ achieves consistently higher scores than its single-call counterpart.
Building on this stronger multi-step setting, \ours$^\dag$~further boosts performance by replacing textified coordinates with continuous coordinate actions, enabling more precise region localization.

\subsection{Results Beyond High-Resolution Benchmarks}
\label{sec:general_results}

\begin{table*}[t]
  \centering
  \begin{minipage}[c]{0.5\textwidth}
    \centering
    {\fontsize{9}{10}\selectfont
    \setlength{\tabcolsep}{2pt}
    \begin{tabular}{lccc}
    \toprule[1pt]
    Model & MMMU-Pro & ChartQA & OCRBenchv2 \\ \midrule
    Qwen2.5-VL-7B & 38.3 & 58.9 & 44.7 \\
    DeepEyes-7B   & 44.6 & 62.5 & 46.1 \\
    \ours~-7B     & \textbf{46.7} & \textbf{65.1} & \textbf{47.6} \\
    \bottomrule[1pt]
    \end{tabular}}
  \end{minipage}
  \hfill
  \begin{minipage}[c]{0.45\textwidth}
    \small
    \caption{Performance comparison on general multimodal benchmarks. \ours~achieves the best performance across MMMU-Pro, ChartQA, and OCRBench v2.}
    \label{tab:mmbench}
  \end{minipage}
\end{table*}
\begin{table*}[t]
\centering
\begin{minipage}[c]{0.5\textwidth}
    \centering
    {\fontsize{9}{10}\selectfont
    \setlength{\tabcolsep}{15pt}
    \begin{tabular}{lcc}
    \toprule[1pt]
    Model & MMLU & GSM8K \\
    \midrule
    Qwen2.5-VL-7B & 70.88 & 73.84 \\
    \ours-7B & 71.07 & 73.54 \\
    \bottomrule[1pt]
    \end{tabular}}
\end{minipage}
\hfill
\begin{minipage}[c]{0.45\textwidth}
    \small
    \caption{Performance on pure-text reasoning benchmarks. \ours~maintains comparable performance to Qwen2.5-VL-7B on MMLU and GSM8K.}
    \label{tab:pure_text}
\end{minipage}
\end{table*}
\begin{table*}[t]
\center
\fontsize{9}{10}\selectfont
\setlength{\tabcolsep}{7pt}
\caption{
Performance comparison on grounding and counting benchmarks. \ours~consistently outperforms strong baselines across RefCOCO series, COCO detection, and counting tasks.
}
\label{tab:grounding_counting}
\begin{tabular}{lcccccc}
\toprule[1pt]
Model & RefCOCO & RefCOCO+ & RefCOCOg & COCO & CountBenchQA & TallyQA \\ 
\midrule
Qwen2.5-VL-7B & 87.1 & 79.7 & 83.0 & 19.5 & 77.4 & 71.2 \\
DeepEyes-7B   & 88.5 & 81.5 & 85.1 & 22.8 & 81.9 & 75.9 \\
\ours~-7B     & \textbf{91.7} & \textbf{85.0} & \textbf{87.4} & \textbf{27.4} & \textbf{86.2} & \textbf{80.5} \\
\bottomrule[1pt]
\end{tabular}
\end{table*}
\begin{table*}[!t] 
\center
\fontsize {9}{10} \selectfont
\setlength{\tabcolsep}{6pt}
\caption{Ablation on coordinate policy (Gaussian ($\ell_2^2$) vs.\ Laplace ($\ell_1$)) and uncertainty parameterization (shared $\sigma/\alpha$ vs.\ per-coordinate $\bm{\sigma}/\bm{\alpha}$). Non-shared $\bm{\sigma}/\bm{\alpha}$: per-coordinate scale (standard deviation) predicted by four heads. Laplace policy consistently performs better than Gaussian policy, while using a shared versus per-coordinate $\bm{\sigma}/\bm{\alpha}$ yields comparable results.
}
\label{table:ablation}
\begin{tabular}{ll|ccc|ccc|ccc}
\toprule[1pt]
\multicolumn{2}{c|}{\multirow{2}{*}{Model}} & \multicolumn{3}{c|}{ $V^*$ Bench} & \multicolumn{3}{c|}{HR Bench 4K} & \multicolumn{3}{c}{HR Bench 8K} \\ \cmidrule{3-11}
& & Attr & Spatial & Overall & FSP & FCP & Overall & FSP & FCP & Overall \\ \midrule

\multirow{2}{*}{SFT} 
& $\ell_2^2$ loss & 83.5 & 76.3 & 80.6 & 81.5 & 53.5 & 67.5 & 76.0 & 54.3 & 65.1\\
& $\ell_1$ loss & 84.3 & 78.9 & 82.2 & 82.0 & 55.3 & 68.6 & 77.3 & 54.5 & 65.9 \\ 
 \midrule
\multirow{4}{*}{RL} 
& Gaussian policy & 88.7 & 85.5 & 87.4 & 90.3 & 58.3 & 74.3 & 85.0 & 58.0 & 71.5 \\
& \quad + Independent $\bm{\sigma}$ & 87.8 & 86.8 & 87.4 & 90.0 & 58.3 & 74.1 & 85.5 & 57.8 & 71.6 \\ 
& Laplace policy & 90.4 & 86.8 & 89.0 & 91.0 & 58.5 & 74.8 & 85.8 & 58.3 & 72.0 \\ 
& \quad + Independent $\bm{\alpha}$ & 89.6 & 86.8 & 88.5 & 90.8 & 58.8 & 74.8 & 86.0 & 58.3 & 72.1 \\  \bottomrule[1pt]
\end{tabular}
\end{table*}


\textbf{General multimodal benchmarks.} To evaluate generality beyond grounding-specific benchmarks, we conduct experiments on general multimodal benchmarks, including MMMU-Pro, ChartQA, and OCRBenchv2. As shown in \cref{tab:mmbench}, \ours~consistently improves performance over both the base VLM and DeepEyes. This suggests that our method does not degrade general reasoning ability, but instead provides complementary gains.

\textbf{Pure-text reasoning benchmarks.} To test whether the coordinate branch affects language reasoning, we evaluate the base Qwen2.5-VL-7B and \ours~on MMLU and GSM8K (with strict-match). \ours~scores 71.07\% versus 70.88\% on MMLU and 73.54\% versus 73.84\% on GSM8K, indicating comparable performance on these pure-text tasks (\cref{tab:pure_text}).

\textbf{Object detection benchmarks.} In \cref{tab:grounding_counting}, we evaluate \ours~on standard grounding benchmarks and grounding-sensitive counting tasks, including the RefCOCO series, COCO detection, CountBenchQA, and TallyQA. For single-object grounding benchmarks, we report accuracy. For COCO, which involves multi-object detection, we report mAP following the standard evaluation protocol. Overall, the results further demonstrate the strong grounding capability of our method.

\subsection{Ablation Studies and Analyses}
\label{sec:ablation}

\begin{figure*}[t]
    \centering
    \input{figure/alpha}\hfill
    \input{figure/lambda}
\end{figure*}

We ablate the coordinate policy and uncertainty parameterization, and analyze uncertainty behavior, the effect of the training coefficient, RL training dynamics, and qualitative grounding.

\textbf{Regression loss and policy distribution.} We jointly analyze the choice of regression loss in SFT and policy distribution in RL, as both stages reflect continuous modeling of coordinates.
In the SFT stage, the $\ell_1$ loss consistently outperforms the $\ell_2^2$ loss across all benchmarks, improving overall accuracy on $V^*$ Bench by $+1.6\%$ and yielding consistent gains on HR-Bench 4K ($+1.1\%$) and 8K ($+0.8\%$).
These trends carry over to RL: the Laplace policy consistently outperforms the Gaussian policy on all benchmarks (\cref{table:ablation}).
This is consistent with prior observations in bounding-box and keypoint regression that $\ell_1$-type objectives are more robust for localization.
Under SFT, we further measure bounding-box IoU on the Vis-CoT-363K validation set.
\cref{tab:IoU} shows that \ours~significantly improves localization accuracy over Vis-CoT, with the $\ell_1$ loss achieving the best IoU.

\textbf{Shared vs. independent $\bm{\sigma/\alpha}$ parameterization.} We further investigate whether the dispersion parameter should be shared across coordinates or predicted independently.
Taking the Gaussian policy as an example, we extend the model with four additional heads
$\Sigma=[\mathbf{w}_{\sigma,x_1}, \mathbf{w}_{\sigma,y_1}, \mathbf{w}_{\sigma,x_2}, \mathbf{w}_{\sigma,y_2}]^\top$
to predict per-coordinate standard deviations.
Specifically, the independent standard deviations are 
\begin{equation}
\bm{\sigma}
=[\sigma_1,\sigma_2,\sigma_3,\sigma_4]^\top=
\max\!\left(\Sigma\mathbf{h}, \varepsilon\right),
\end{equation}
which generalizes the shared-variance formulation in Eq.~(\ref{eq:std}).
Now, the importance ratio changes to:
\begin{equation}
r(\theta)
=
\prod_{j=1}^{4}
\frac{\sigma_{\theta_{\mathsf{old}},j}}{\sigma_{\theta,j}}
\exp\!\left(
-\frac{(b_{j}-\mu_{\theta,j})^2}{2\sigma_{\theta,j}^2}
+
\frac{(b_{j}-\mu_{\theta_{\mathsf{old}},j})^2}{2\sigma_{\theta_{\mathsf{old}},j}^2}
\right).
\end{equation}
The Laplace policy follows an analogous procedure.
As shown in \cref{table:ablation}, the performance gap between shared and independent parameterizations is marginal across all benchmarks.
Given the comparable results, we adopt a shared $\sigma/\alpha$ in our final model for simplicity and efficiency.

\textbf{Behavior analysis of $\alpha$.} In \cref{fig:alpha}, we  analyze the behavior of the Laplace scale parameter $\alpha$ by comparing its average values across success and failure cases. Results across different image resolutions are normalized. 
Successful trajectories consistently exhibit smaller $\alpha$ than failed ones. This observation indicates that accurate localization is associated with lower predictive uncertainty, leading the policy to produce more concentrated bounding-box samples.

\textbf{Effect of $\lambda$.} In \cref{fig:lambda}, we study the effect of $\lambda$ in Eq.~(\ref{eq:sft}) by varying it from $0.1$ to $0.9$ and reporting overall accuracy on three benchmarks.
\ours~achieves the best performance at $\lambda=0.3$, while maintaining relatively stable accuracy across a wide range of values.
Notably, across all $\lambda$ settings, \ours~consistently outperforms the text-based baselines by a clear margin, indicating that the performance gains are robust and not sensitive to the choice of $\lambda$.

\begin{figure*}[t]
\centering
\begin{minipage}[t]{0.48\textwidth}
\vspace{0pt}
\centering
\includegraphics[width=\linewidth]{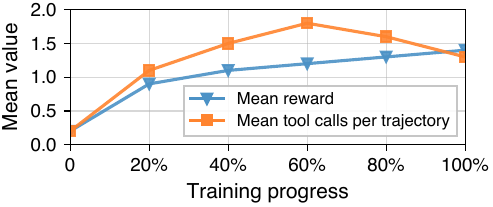}
\captionof{figure}{RL training dynamics. Mean reward increases throughout training, while tool calls per trajectory peak at 60\% and decline thereafter.}
\label{fig:rl_training_dynamics}
\end{minipage}\hfill
\begin{minipage}[t]{0.48\textwidth}
\vspace{0pt}
\centering
\fontsize{9}{10}\selectfont
\setlength{\tabcolsep}{8pt}
\captionof{table}{RL policy statistics for text and coordinate actions. Clipping uses $\varepsilon=0.2$; gradient-norm shares are normalized across branches.}
\label{tab:action_balance}
\begin{tabular}{lcc}
\toprule
Action type & \makecell{Text\\tokens} & \makecell{Coordinate\\actions} \\
\midrule
Mean ratio & 1.012 & 1.018 \\
Std. ratio & 0.071 & 0.093 \\
Clipped fraction & 4.6\% & 6.8\% \\
Gradient share & 54\% & 46\% \\
\bottomrule
\end{tabular}

\end{minipage}
\end{figure*}

\textbf{RL training diagnostics.} To characterize training progress and assess whether either action type dominates optimization, Figure~\ref{fig:rl_training_dynamics} tracks mean reward and tool calls per trajectory, which shows that mean reward rises from 0.2 to 1.4, while tool calls peak at 1.8 per trajectory at 60\% of training before declining to 1.3. Table~\ref{tab:action_balance} compares their optimization statistics, which shows similar importance-ratio means and clipping fractions for text and coordinate actions, with normalized gradient-norm contributions of 54\% and 46\%, respectively. Together, these results show no steadily increasing tool use alongside reward gains and no marked optimization imbalance between action types.

\begin{figure*}[t]
    \centering
    \includegraphics[width=1\linewidth]{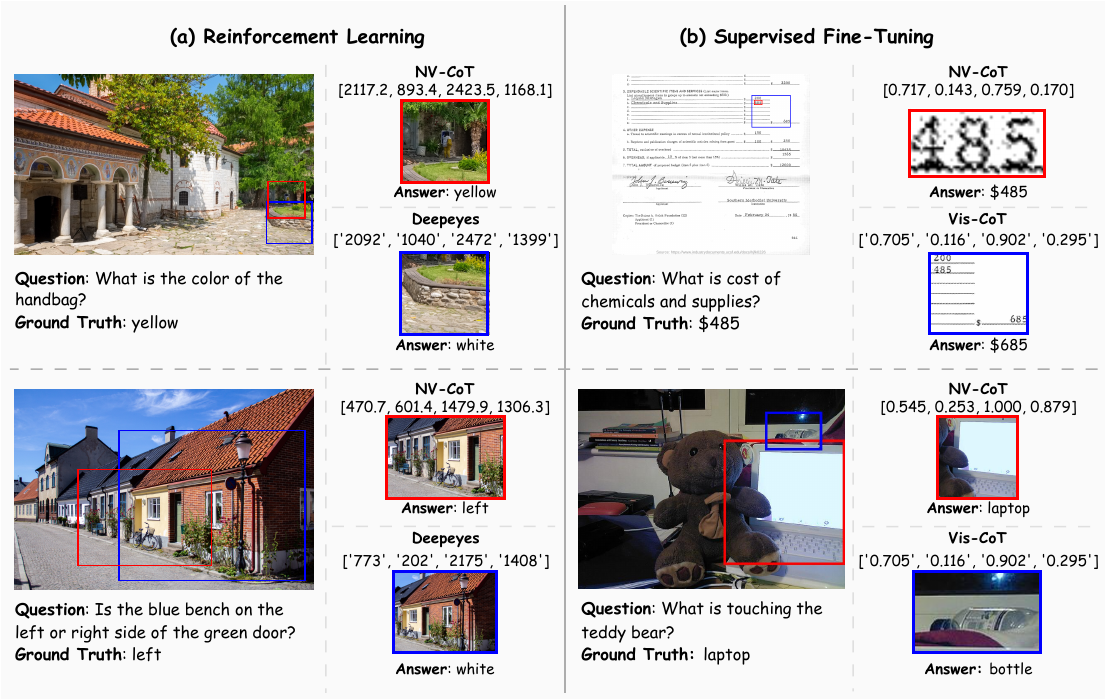}
    \caption{
Visualization of bounding boxes. \ours~produces more accurate bounding boxes (shown in {\color{red} \textbf{red}}) compared to the backbone model (shown in {\color{blue} \textbf{blue}}), showing improved localization capability.}
    \label{fig:visualization}
    \vspace{-1mm}
\end{figure*}

\textbf{Qualitative grounding visualization.} In \cref{fig:visualization}, we compare bounding-box localization between \ours~and the backbone model. The backbone often produces coarse or loosely aligned boxes that include excess background or miss parts of target objects. In contrast, \ours~produces tighter boxes that better align with target extents, illustrating improved localization.

\section{Conclusion and limitation}
\label{sec:conclusion}

We present \ours, a plug-and-play framework that enables MLLMs to directly generate continuous bounding-box coordinates instead of textual tokens. By formulating localization as a continuous-action problem with Gaussian/Laplace policies, \ours~supports both supervised fine-tuning and reinforcement learning. \ours~mitigates the modality mismatch and semantic fragmentation of text-based visual CoT, while avoiding the fixed-granularity limitation of patch-based designs, with only minimal architectural modifications. Experiments across high-resolution, general multimodal, grounding, and counting benchmarks show consistent gains in visual reasoning and localization, while performance on pure-text tasks remains comparable to the base model. These gains are accompanied by improved final-answer accuracy and faster convergence.

  We acknowledge that \ours~does not address cases where the underlying MLLM fails to correctly understand the image or the task, as our method focuses on improving visual grounding capability. However, in scenarios where the model has a reasonable understanding but produces inaccurate localization, \ours~consistently improves precision through continuous coordinate prediction.
\clearpage
\section*{Acknowledgements and Disclosures}

\textbf{Funding.}
This research is supported by the National Research Foundation, Singapore under its AI Singapore Programme (AISG Award No: AISG3-RP-2022-030). This research is also supported by National Research Foundation, Singapore, NRF-NRFI10-2024-0004.

\textbf{Competing Interests.}
The authors declare no competing interests.

\bibliography{main}

@inproceedings{zheng2025deepeyes,
  title={DeepEyes: Incentivizing ``Thinking with Images'' via Reinforcement Learning},
  author={Zheng, Ziwei and Yang, Minghao and Hong, Jack and Zhao, Chenxiao and Xu, Guohai and Yang, Le and Shen, Chao and Yu, Xing},
  booktitle={ICLR},
  year={2026}
}

@inproceedings{shao2024visual,
  title={Visual cot: Advancing multi-modal language models with a comprehensive dataset and benchmark for chain-of-thought reasoning},
  author={Shao, Hao and Qian, Shengju and Xiao, Han and Song, Guanglu and Zong, Zhuofan and Wang, Letian and Liu, Yu and Li, Hongsheng},
  booktitle={NeurIPS},
  year={2024}
}

@inproceedings{zhao2025unsupervised,
  title={Unsupervised visual chain-of-thought reasoning via preference optimization},
  author={Zhao, Kesen and Zhu, Beier and Sun, Qianru and Zhang, Hanwang},
  booktitle={ICCV},
  year={2025}
}

@inproceedings{
li2025imagine,
title={Imagine While Reasoning in Space: Multimodal Visualization-of-Thought},
author={Chengzu Li and Wenshan Wu and Huanyu Zhang and Yan Xia and Shaoguang Mao and Li Dong and Ivan Vuli{\'c} and Furu Wei},
booktitle={ICML},
year={2025}
}

@article{su2025openthinkimg,
  title={Openthinkimg: Learning to think with images via visual tool reinforcement learning},
  author={Su, Zhaochen and Li, Linjie and Song, Mingyang and Hao, Yunzhuo and Yang, Zhengyuan and Zhang, Jun and Chen, Guanjie and Gu, Jiawei and Li, Juntao and Qu, Xiaoye and others},
  journal={arXiv preprint arXiv:2505.08617},
  year={2025}
}

@inproceedings{liu2025visionreasoner,
  title={VisionReasoner: Unified Reasoning-Integrated Visual Perception via Reinforcement Learning},
  author={Liu, Yuqi and Qu, Tianyuan and Zhong, Zhisheng and Peng, Bohao and Liu, Shu and Yu, Bei and Jia, Jiaya},
  booktitle={ICLR},
  year={2026}
}

@article{bai2025qwen2,
  title={Qwen2.5-vl technical report},
  author={Bai, Shuai and Chen, Keqin and Liu, Xuejing and Wang, Jialin and Ge, Wenbin and Song, Sibo and Dang, Kai and Wang, Peng and Wang, Shijie and Tang, Jun and others},
  journal={arXiv preprint arXiv:2502.13923},
  year={2025}
}

@article{zhu2025internvl3,
  title={Internvl3: Exploring advanced training and test-time recipes for open-source multimodal models},
  author={Zhu, Jinguo and Wang, Weiyun and Chen, Zhe and Liu, Zhaoyang and Ye, Shenglong and Gu, Lixin and Tian, Hao and Duan, Yuchen and Su, Weijie and Shao, Jie and others},
  journal={arXiv preprint arXiv:2504.10479},
  year={2025}
}

@inproceedings{li2025latent,
  title={Latent Visual Reasoning},
  author={Li, Bangzheng and Sun, Ximeng and Liu, Jiang and Wang, Ze and Wu, Jialian and Yu, Xiaodong and Barsoum, Emad and Chen, Muhao and Liu, Zicheng},
  booktitle={ICLR},
  year={2026}
}

@inproceedings{su2025patch,
  title={Patch-as-Decodable-Token: Towards Unified Multi-Modal Vision Tasks in MLLMs},
  author={Su, Yongyi and Zhang, Haojie and Li, Shijie and Liu, Nanqing and Liao, Jingyi and Pan, Junyi and Liu, Yuan and Xing, Xiaofen and Sun, Chong and Li, Chen and Chen, Nancy and Yan, Shuicheng and Yang, Xulei and Xu, Xun},
  booktitle={ICLR},
  year={2026}
}

@article{li2024llava,
  title={Llava-onevision: Easy visual task transfer},
  author={Li, Bo and Zhang, Yuanhan and Guo, Dong and Zhang, Renrui and Li, Feng and Zhang, Hao and Zhang, Kaichen and Zhang, Peiyuan and Li, Yanwei and Liu, Ziwei and others},
  journal={TMLR},
  year={2025}
}

@inproceedings{wu2024v,
  title={V*: Guided visual search as a core mechanism in multimodal llms},
  author={Wu, Penghao and Xie, Saining},
  booktitle={CVPR},
  year={2024}
}

@inproceedings{wang2025divide,
  title={Divide, conquer and combine: A training-free framework for high-resolution image perception in multimodal large language models},
  author={Wang, Wenbin and Ding, Liang and Zeng, Minyan and Zhou, Xiabin and Shen, Li and Luo, Yong and Yu, Wei and Tao, Dacheng},
  booktitle={AAAI},
  year={2025}
}

@inproceedings{li2023blip,
  title={Blip-2: Bootstrapping language-image pre-training with frozen image encoders and large language models},
  author={Li, Junnan and Li, Dongxu and Savarese, Silvio and Hoi, Steven},
  booktitle={ICML},
  year={2023}
}

@inproceedings{liu2023visual,
  title={Visual instruction tuning},
  author={Liu, Haotian and Li, Chunyuan and Wu, Qingyang and Lee, Yong Jae},
  booktitle={NeurIPS},
  year={2023}
}

@inproceedings{liu2024improved,
  title={Improved baselines with visual instruction tuning},
  author={Liu, Haotian and Li, Chunyuan and Li, Yuheng and Lee, Yong Jae},
  booktitle={CVPR},
  year={2024}
}

@inproceedings{liu2024llava,
  title={Llava-plus: Learning to use tools for creating multimodal agents},
  author={Liu, Shilong and Cheng, Hao and Liu, Haotian and Zhang, Hao and Li, Feng and Ren, Tianhe and Zou, Xueyan and Yang, Jianwei and Su, Hang and Zhu, Jun and others},
  booktitle={ECCV},
  year={2024}
}

@inproceedings{guo2024llava,
  title={Llava-uhd: an lmm perceiving any aspect ratio and high-resolution images},
  author={Guo, Zonghao and Xu, Ruyi and Yao, Yuan and Cui, Junbo and Ni, Zanlin and Ge, Chunjiang and Chua, Tat-Seng and Liu, Zhiyuan and Huang, Gao},
  booktitle={ECCV},
  year={2024}
}

@inproceedings{zhang2025llava,
  title={LLaVA-Mini: Efficient Image and Video Large Multimodal Models with One Vision Token},
  author={Zhang, Shaolei and Fang, Qingkai and Yang, Zhe and Feng, Yang},
  booktitle={ICLR},
  year={2025}
}

@inproceedings{lin2024video,
  title={Video-llava: Learning united visual representation by alignment before projection},
  author={Lin, Bin and Ye, Yang and Zhu, Bin and Cui, Jiaxi and Ning, Munan and Jin, Peng and Yuan, Li},
  booktitle={EMNLP},
  year={2024}
}

@inproceedings{li2023llava,
  title={Llava-med: Training a large language-and-vision assistant for biomedicine in one day},
  author={Li, Chunyuan and Wong, Cliff and Zhang, Sheng and Usuyama, Naoto and Liu, Haotian and Yang, Jianwei and Naumann, Tristan and Poon, Hoifung and Gao, Jianfeng},
  booktitle={NeurIPS},
  year={2023}
}

@article{bai2023qwen,
  title={Qwen technical report},
  author={Bai, Jinze and Bai, Shuai and Chu, Yunfei and Cui, Zeyu and Dang, Kai and Deng, Xiaodong and Fan, Yang and Ge, Wenbin and Han, Yu and Huang, Fei and others},
  journal={arXiv preprint arXiv:2309.16609},
  year={2023}
}

@article{wang2024qwen2,
  title={Qwen2-vl: Enhancing vision-language model's perception of the world at any resolution},
  author={Wang, Peng and Bai, Shuai and Tan, Sinan and Wang, Shijie and Fan, Zhihao and Bai, Jinze and Chen, Keqin and Liu, Xuejing and Wang, Jialin and Ge, Wenbin and others},
  journal={arXiv preprint arXiv:2409.12191},
  year={2024}
}

@inproceedings{chen2024internvl,
  title={Internvl: Scaling up vision foundation models and aligning for generic visual-linguistic tasks},
  author={Chen, Zhe and Wu, Jiannan and Wang, Wenhai and Su, Weijie and Chen, Guo and Xing, Sen and Zhong, Muyan and Zhang, Qinglong and Zhu, Xizhou and Lu, Lewei and others},
  booktitle={CVPR},
  year={2024}
}

@article{lu2025internvl,
  title={Internvl-x: Advancing and accelerating internvl series with efficient visual token compression},
  author={Lu, Dongchen and Sun, Yuyao and Zhang, Zilu and Huang, Leping and Zeng, Jianliang and Shu, Mao and Cao, Huo},
  journal={arXiv preprint arXiv:2503.21307},
  year={2025}
}

@inproceedings{wang2025pixel,
  title={Pixel reasoner: Incentivizing pixel-space reasoning with curiosity-driven reinforcement learning},
  author={Wang, Haozhe and Su, Alex and Ren, Weiming and Lin, Fangzhen and Chen, Wenhu},
  booktitle={NeurIPS},
  year={2025}
}

@article{zhang2025chain,
  title={Chain-of-Focus: Adaptive Visual Search and Zooming for Multimodal Reasoning via RL},
  author={Zhang, Xintong and Gao, Zhi and Zhang, Bofei and Li, Pengxiang and Zhang, Xiaowen and Liu, Yang and Yuan, Tao and Wu, Yuwei and Jia, Yunde and Zhu, Song-Chun and others},
  journal={arXiv preprint arXiv:2505.15436},
  year={2025}
}

@inproceedings{huang2025visualtoolagent,
  title={Learning to Select Visual Tools from Experience},
  author={Huang, Zeyi and Ji, Yuyang and Rajan, Anirudh Sundara and Cai, Zefan and Xiao, Wen and Wang, Haohan and Hu, Junjie and Lee, Yong Jae},
  booktitle={CVPR},
  year={2026}
}

@inproceedings{wang2025vgr,
  title={VGR: Visual Grounded Reasoning},
  author={Wang, Jiacong and Kang, Zijian and Wang, Haochen and Xiao, Liang and Wang, Ya and Li, Jiawen and Wu, Bohong and Ran, Jiao and Jiang, Haiyong and Feng, Chao and Xiao, Jun},
  booktitle={ICLR},
  year={2026}
}

@article{chung2025don,
  title={Don't Look Only Once: Towards Multimodal Interactive Reasoning with Selective Visual Revisitation},
  author={Chung, Jiwan and Kim, Junhyeok and Kim, Siyeol and Lee, Jaeyoung and Kim, Min Soo and Yu, Youngjae},
  journal={arXiv preprint arXiv:2505.18842},
  year={2025}
}

@inproceedings{zhang2025cmmcot,
  title={CMMCoT: Enhancing Complex Multi-Image Comprehension via Multi-Modal Chain-of-Thought and Memory Augmentation},
  author={Zhang, Guanghao and Zhong, Tao and Xia, Yan and Liu, Mushui and Yu, Zhelun and Li, Haoyuan and He, Wanggui and She, Dong and Wang, Yi and Jiang, Hao},
  booktitle={AAAI},
  year={2026}
}

@inproceedings{shao2024deepseekmathpushinglimitsmathematical,
  title={DeepSeekMath: Pushing the Limits of Mathematical Reasoning in Open Language Models},
  author={Shao, Zhihong and Wang, Peiyi and Zhu, Qihao and Xu, Runxin and Song, Junxiao and Bi, Xiao and Zhang, Haowei and Zhang, Mingchuan and Li, Y. K. and Wu, Y. and Guo, Daya},
  booktitle={ICLR},
  year={2025}
}

@InProceedings{Girshick_2015_ICCV,
author = {Girshick, Ross},
title = {Fast R-CNN},
booktitle = {ICCV},
year = {2015}
}

@inproceedings{sun2018integral,
  title={Integral human pose regression},
  author={Sun, Xiao and Xiao, Bin and Wei, Fangyin and Liang, Shuang and Wei, Yichen},
  booktitle={ECCV},
  year={2018}
}

@inproceedings{mathew2021docvqa,
  title={Docvqa: A dataset for vqa on document images},
  author={Mathew, Minesh and Karatzas, Dimosthenis and Jawahar, CV},
  booktitle={WACV},
  year={2021}
}

@inproceedings{singh2019towards,
  title={Towards vqa models that can read},
  author={Singh, Amanpreet and Natarajan, Vivek and Shah, Meet and Jiang, Yu and Chen, Xinlei and Batra, Dhruv and Parikh, Devi and Rohrbach, Marcus},
  booktitle={CVPR},
  year={2019}
}

@inproceedings{van2023document,
  title={Document understanding dataset and evaluation (dude)},
  author={Van Landeghem, Jordy and Tito, Rub{\`e}n and Borchmann, {\L}ukasz and Pietruszka, Micha{\l} and Joziak, Pawel and Powalski, Rafal and Jurkiewicz, Dawid and Coustaty, Micka{\"e}l and Anckaert, Bertrand and Valveny, Ernest and others},
  booktitle={ICCV},
  year={2023}
}

@inproceedings{huang2019icdar2019,
  title={Icdar2019 competition on scanned receipt ocr and information extraction},
  author={Huang, Zheng and Chen, Kai and He, Jianhua and Bai, Xiang and Karatzas, Dimosthenis and Lu, Shijian and Jawahar, CV},
  booktitle={ICDAR},
  year={2019}
}

@inproceedings{plummer2015flickr30k,
  title={Flickr30k entities: Collecting region-to-phrase correspondences for richer image-to-sentence models},
  author={Plummer, Bryan A and Wang, Liwei and Cervantes, Chris M and Caicedo, Juan C and Hockenmaier, Julia and Lazebnik, Svetlana},
  booktitle={ICCV},
  year={2015}
}

@inproceedings{zhu2016visual7w,
  title={Visual7w: Grounded question answering in images},
  author={Zhu, Yuke and Groth, Oliver and Bernstein, Michael and Fei-Fei, Li},
  booktitle={CVPR},
  year={2016}
}

@inproceedings{mathew2022infographicvqa,
  title={Infographicvqa},
  author={Mathew, Minesh and Bagal, Viraj and Tito, Rub{\`e}n and Karatzas, Dimosthenis and Valveny, Ernest and Jawahar, CV},
  booktitle={WACV},
  year={2022}
}

@inproceedings{hudson2019gqa,
  title={Gqa: A new dataset for real-world visual reasoning and compositional question answering},
  author={Hudson, Drew A and Manning, Christopher D},
  booktitle={CVPR},
  year={2019}
}

@inproceedings{li2024multimodal,
  title={Multimodal {A}r{X}iv: A Dataset for Improving Scientific Comprehension of Large Vision-Language Models},
  author={Li, Lei and Wang, Yuqi and Xu, Runxin and Wang, Peiyi and Feng, Xiachong and Kong, Lingpeng and Liu, Qi},
  booktitle={ACL},
  year={2024}
}

@inproceedings{wang2025sota,
  title={SoTA with Less: MCTS-Guided Sample Selection for Data-Efficient Visual Reasoning Self-Improvement},
  author={Wang, Xiyao and Yang, Zhengyuan and Feng, Chao and Lu, Hongjin and Li, Linjie and Lin, Chung-Ching and Lin, Kevin and Huang, Furong and Wang, Lijuan},
  booktitle={NeurIPS},
  year={2025}
}

@inproceedings{shen2025zoomeye,
  title={Zoomeye: Enhancing multimodal llms with human-like zooming capabilities through tree-based image exploration},
  author={Shen, Haozhan and Zhao, Kangjia and Zhao, Tiancheng and Xu, Ruochen and Zhang, Zilun and Zhu, Mingwei and Yin, Jianwei},
  booktitle={EMNLP},
  year={2025}
}

@inproceedings{wang2025retrieval,
  title={Retrieval-Augmented Perception: High-Resolution Image Perception Meets Visual {RAG}},
  author={Wang, Wenbin and Jing, Yongcheng and Ding, Liang and Wang, Yingjie and Shen, Li and Luo, Yong and Du, Bo and Tao, Dacheng},
  booktitle={ICML},
  year={2025}
}

@inproceedings{li2025dyfo,
  title={Dyfo: A training-free dynamic focus visual search for enhancing lmms in fine-grained visual understanding},
  author={Li, Geng and Xu, Jinglin and Zhao, Yunzhen and Peng, Yuxin},
  booktitle={CVPR},
  year={2025}
}

@inproceedings{hendrycks2021measuring,
  title={Measuring Massive Multitask Language Understanding},
  author={Hendrycks, Dan and Burns, Collin and Basart, Steven and Zou, Andy and Mazeika, Mantas and Song, Dawn and Steinhardt, Jacob},
  booktitle={ICLR},
  year={2021}
}

@article{cobbe2021training,
  title={Training Verifiers to Solve Math Word Problems},
  author={Cobbe, Karl and Kosaraju, Vineet and Bavarian, Mohammad and Chen, Mark and Jun, Heewoo and Kaiser, Lukasz and Plappert, Matthias and Tworek, Jerry and Hilton, Jacob and Nakano, Reiichiro and Hesse, Christopher and Schulman, John},
  journal={arXiv preprint arXiv:2110.14168},
  year={2021}
}

@inproceedings{yue2024mmmu_pro,
  title={{MMMU}-Pro: A More Robust Multi-discipline Multimodal Understanding Benchmark},
  author={Yue, Xiang and Zheng, Tianyu and Ni, Yuansheng and Wang, Yubo and Zhang, Kai and Tong, Shengbang and Sun, Yuxuan and Yu, Botao and Zhang, Ge and Sun, Huan and Su, Yu and Chen, Wenhu and Neubig, Graham},
  booktitle={ACL},
  year={2025}
}

@inproceedings{masry2022chartqa,
  title={ChartQA: A Benchmark for Question Answering about Charts with Visual and Logical Reasoning},
  author={Masry, Ahmed and Long, Do Xuan and Tan, Jia Qing and Joty, Shafiq and Hoque, Enamul},
  booktitle={ACL Findings},
  year={2022}
}

@inproceedings{fu2025ocrbenchv2,
  title={OCRBench v2: An Improved Benchmark for Evaluating Large Multimodal Models on Visual Text Localization and Reasoning},
  author={Fu, Ling and Kuang, Zhebin and Song, Jiajun and Huang, Mingxin and Yang, Biao and Li, Yuzhe and Zhu, Linghao and Luo, Qidi and Wang, Xinyu and Lu, Hao and Li, Zhang and Tang, Guozhi and Shan, Bin and Lin, Chunhui and Liu, Qi and Wu, Binghong and Feng, Hao and Liu, Hao and Huang, Can and Tang, Jingqun and Chen, Wei and Jin, Lianwen and Liu, Yuliang and Bai, Xiang},
  booktitle={NeurIPS},
  year={2025}
}

@inproceedings{yu2016modeling,
  title={Modeling Context in Referring Expressions},
  author={Yu, Licheng and Poirson, Patrick and Yang, Shan and Berg, Alexander C. and Berg, Tamara L.},
  booktitle={ECCV},
  year={2016}
}

@inproceedings{mao2016generation,
  title={Generation and Comprehension of Unambiguous Object Descriptions},
  author={Mao, Junhua and Huang, Jonathan and Toshev, Alexander and Camburu, Oana and Yuille, Alan L. and Murphy, Kevin},
  booktitle={CVPR},
  year={2016}
}

@inproceedings{lin2014coco,
  title={{Microsoft COCO: Common Objects in Context}},
  author={Lin, Tsung-Yi and Maire, Michael and Belongie, Serge and Bourdev, Lubomir and Girshick, Ross and Hays, James and Perona, Pietro and Ramanan, Deva and Doll{\'a}r, Piotr and Zitnick, C. Lawrence},
  booktitle={ECCV},
  year={2014}
}

@article{beyer2024paligemma,
  title={PaliGemma: A versatile 3B VLM for transfer},
  author={Beyer, Lucas and Steiner, Andreas and Susano Pinto, Andr{\'e} and Kolesnikov, Alexander and Wang, Xiao and Salz, Daniel and Neumann, Maxim and Alabdulmohsin, Ibrahim and Tschannen, Michael and Bugliarello, Emanuele and Unterthiner, Thomas and Keysers, Daniel and Koppula, Skanda and Liu, Fangyu and Grycner, Adam and Gritsenko, Alexey and Houlsby, Neil and Kumar, Manoj and Rong, Keran and Eisenschlos, Julian and Kabra, Rishabh and Bauer, Matthias and Bo\v{s}njak, Matko and Chen, Xi and Minderer, Matthias and Voigtlaender, Paul and Bica, Ioana and Balazevic, Ivana and Puigcerver, Joan and Papalampidi, Pinelopi and Henaff, Olivier and Xiong, Xi and Soricut, Radu and Harmsen, Jeremiah and Zhai, Xiaohua},
  journal={arXiv preprint arXiv:2407.07726},
  year={2024}
}

@inproceedings{acharya2019tallyqa,
  title={TallyQA: Answering Complex Counting Questions},
  author={Acharya, Manoj and Kafle, Kushal and Kanan, Christopher},
  booktitle={AAAI},
  year={2019}
}
\bibliographystyle{unsrtnat}

\appendix
\section{Implementation Details}
\label{sec:implementation_details}

\subsection{Datasets}
\label{sec:datasets}

For fair comparison with prior visual chain-of-thought methods, we adopt exactly the same training data as the corresponding backbone models.

\textbf{Supervised Fine-Tuning (SFT).}
For supervised fine-tuning, we adopt the Visual-CoT dataset constructed following the protocol of Vis-CoT~\cite{shao2024visual}. The dataset is built by collecting images and annotations from twelve publicly available datasets spanning five domains, including text/document understanding, fine-grained visual recognition, general visual question answering, chart understanding, and relation reasoning. These source datasets mainly consist of VQA and image captioning benchmarks, from which images and useful annotations such as question-answer pairs, captions, object bounding boxes, and object relations are reused.
The data construction process involves both linguistic and visual annotation stages. For linguistic annotation, a large language model (GPT-4) is employed to generate or refine question-answer pairs and, where applicable, to produce intermediate reasoning steps. For visual annotation, task-relevant regions are identified to serve as visual chain-of-thought (CoT) grounding signals. In text-centric domains, such as TextVQA~\cite{singh2019towards}, DocVQA~\cite{mathew2021docvqa}, DUDE~\cite{van2023document}, SROIE~\cite{huang2019icdar2019}, and InfographicsVQA~\cite{mathew2022infographicvqa}, OCR-based tools (e.g., PaddleOCR) are used to detect text regions, and the CoT bounding boxes are defined as regions containing words or sentences that directly support the answer. In fine-grained recognition tasks, bounding boxes corresponding to discriminative object parts or attributes are used to guide fine-grained visual reasoning.
For general VQA and relation reasoning tasks, including Flickr30k~\cite{plummer2015flickr30k}, Visual7W~\cite{zhu2016visual7w}, VSR~\cite{liu2023visual}, and GQA~\cite{hudson2019gqa}, object-level annotations and scene graph information are leveraged. The visual CoT bounding boxes correspond to objects or regions that are explicitly involved in answering the question, such as spatially related objects or attributes specified in the query. In particular, for GQA, structured semantic annotations are used to identify the target objects and relations, enabling precise grounding of intermediate reasoning steps.

During SFT, each training example consists of an image, a question, a sequence of intermediate visual grounding steps represented by bounding boxes, and the final answer. The model is trained to jointly generate textual outputs and predict continuous bounding-box coordinates, allowing it to learn region-grounded visual reasoning under explicit supervision.

\textbf{Reinforcement Learning (RL).}
For reinforcement learning, we adopt the same training data and protocol as DeepEyes~\cite{zheng2025deepeyes}. Unlike supervised fine-tuning, the RL setting does not rely on any explicit intermediate bounding-box annotations. Instead, the model is optimized solely through outcome-level supervision, which better reflects practical scenarios where fine-grained visual grounding labels are unavailable.

The RL training data is constructed to emphasize tool-aware visual reasoning and consists of three complementary subsets: fine-grained visual search data, chart understanding data, and general reasoning data. The fine-grained subset is sampled from a portion of the V$^*$ training set~\cite{wu2024v}, focusing on high-resolution images and detailed perception tasks where localized visual inspection is critical for correct answering. The chart subset is drawn from ArxivQA~\cite{li2024multimodal}, which contains synthetic plots, diagrams, and structured graphical representations, significantly expanding the diversity of visual modalities beyond natural images. To further enhance reasoning diversity, some samples are incorporate from ThinkLite-VL~\cite{wang2025sota}, which include multimodal questions requiring arithmetic reasoning, commonsense inference, and multi-step problem solving.
Then a dedicated data selection pipeline is applied. First, question difficulty is estimated by prompting a strong vision-language model to generate multiple responses per question and measuring answer consistency; overly trivial or extremely difficult samples are filtered out. Second, original questions are reformulated into open-ended formats to better align with generative training objectives, and samples that cannot be reliably converted are discarded. Third, unverifiable instances are removed, including those with ambiguous questions, incorrect annotations, or unreadable visual content. Finally, chart data is exempted from this filtering step due to its inherent reliance on visual structure, while reasoning data is retained in its original form.

\subsection{Evaluation Protocol}
\label{sec:evaluation}

\textbf{Inference Prompts.}
We employ two different prompt templates for reinforcement learning and supervised fine-tuning, respectively, reflecting their distinct training objectives and supervision signals. For reinforcement learning, we use a structured inference prompt that explicitly encourages step-by-step reasoning and conditional tool usage. The prompt instructs the model to first reason about the question, then optionally invoke the image zoom-in tool when fine-grained visual inspection is necessary, and finally produce the answer in a strictly defined format. This structured output separates reasoning, tool calls, and final answers, enabling reliable trajectory parsing and outcome-driven reward computation during policy optimization.
For supervised fine-tuning, we adopt a localization-oriented prompt that explicitly asks the model to identify a task-relevant image region by predicting a bounding-box coordinate. The model is instructed to answer the question based on both the original image and the localized region. This prompt design encourages the model to learn explicit visual grounding under bounding-box supervision, strengthening the alignment between region selection and downstream reasoning.

\begin{tcolorbox}[
    colback=gray!20, 
    colframe=black!70, 
    arc=2mm, 
    auto outer arc, 
    boxrule=0.3mm, 
    width=0.97\textwidth, 
    title=\centering RL Evaluating Prompt Template
]
\textit{
Think first, call image\_zoom\_in\_tool if needed, then answer. \\
Format strictly as: \\
\texttt{\textless think\textgreater...\textless /think\textgreater } \\
\texttt{\textless tool\_call\textgreater...\textless /tool\_call\textgreater }~(if tools needed)  \\
\texttt{\textless answer\textgreater...\textless /answer\textgreater }
}
\end{tcolorbox}

\begin{tcolorbox}[
    colback=gray!20, 
    colframe=black!70, 
    arc=2mm, 
    auto outer arc, 
    boxrule=0.3mm, 
    width=0.97\textwidth, 
    title=\centering SFT Evaluating Prompt Template
]
\textit{
Please provide the bounding box coordinate of the region that can help you answer the question better. \\
Answer the question based on the original image and local image.
}
\end{tcolorbox}

\subsection{{Coordinate and Crop Handling}}
{For iterative zoom-in, predicted bounding boxes use the original image's global coordinate frame. Before cropping, normalized coordinates are converted to pixel coordinates when needed and adjusted for image padding; the final crop window is clipped to the image boundaries. Coordinates with invalid corner ordering are not reordered: the crop is formed around the averaged center with a minimum crop size and then clipped. Invalid coordinate predictions occur more often early in RL training and decrease rapidly; none were observed during inference.}

\subsection{More Experimental Results}
\label{sec:More_results}

\begin{figure*}[th]
    \centering
    \includegraphics[width=1\linewidth]{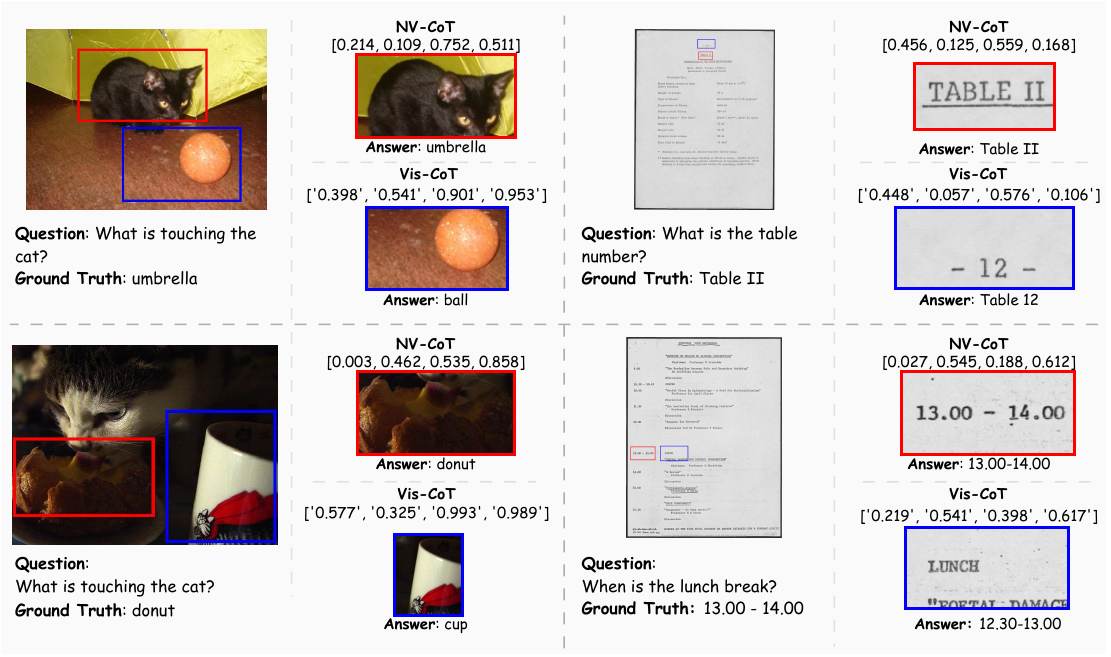}
    \caption{
More visualization results of bounding boxes. \ours~produces more accurate bounding boxes (shown in {\color{red} \textbf{red}}) compared to the backbone model (shown in {\color{blue} \textbf{blue}}), demonstrating improved localization capability.}
    \label{fig:SFT_vis}
    \vspace{-1mm}
\end{figure*}

\textbf{Additional visualization.} In \cref{fig:SFT_vis}, we provide additional qualitative examples of bounding-box localization. As in the main-text visualization, the backbone model (blue) tends to produce coarse or loosely aligned boxes, while \ours~(red) more closely matches object extents, illustrating improved localization quality.




\section{Impact Statements}
\label{impact}
This paper presents work whose primary goal is to advance the capabilities of multimodal large language models in visual reasoning, particularly through more accurate and efficient region grounding. By enabling models to reason over images using continuous numerical representations, our approach improves localization precision and training efficiency, which may benefit a wide range of downstream applications such as visual question answering and optical character recognition.
At the same time, improved visual localization and reasoning capabilities could be applied in sensitive domains. However, our work does not introduce new data sources, deployable systems, or application-specific pipelines, and it relies on standard benchmark datasets commonly used in the research community. As such, we do not believe that this work introduces novel ethical risks beyond those already well recognized in the field of MLLMs.


\clearpage
\section*{NeurIPS Paper Checklist}

\begin{enumerate}

\item {\bf Claims}
    \item[] Question: Do the main claims made in the abstract and introduction accurately reflect the paper's contributions and scope?
    \item[] Answer: \answerYes{} 
    \item[] Justification: The abstract and introduction describe continuous numerical box prediction for visual CoT, and \S\S\ref{sec:method} and~\ref{sec:experiments} evaluate the proposed SFT and RL variants on the stated benchmarks.

\item {\bf Limitations}
    \item[] Question: Does the paper discuss the limitations of the work performed by the authors?
    \item[] Answer: \answerYes{} 
    \item[] Justification: \S\ref{sec:conclusion} discusses the limitations of the work.

\item {\bf Theory assumptions and proofs}
    \item[] Question: For each theoretical result, does the paper provide the full set of assumptions and a complete (and correct) proof?
    \item[] Answer: \answerNA{} 
    \item[] Justification: The paper presents method derivations and experimental results, but no theorem or formal theoretical result requiring proof.

    \item {\bf Experimental result reproducibility}
    \item[] Question: Does the paper fully disclose all the information needed to reproduce the main experimental results of the paper to the extent that it affects the main claims and/or conclusions of the paper (regardless of whether the code and data are provided or not)?
    \item[] Answer: \answerYes{} 
    \item[] Justification: \S\S\ref{sec:method} and~\ref{sec:setup} describe the method and training configuration; the appendix provides dataset, evaluation-protocol, and prompt details.

\item {\bf Open access to data and code}
    \item[] Question: Does the paper provide open access to the data and code, with sufficient instructions to faithfully reproduce the main experimental results, as described in supplemental material?
    \item[] Answer: \answerYes{} 
    \item[] Justification: The project page links to the public \ours~code repository. The paper and appendix describe the method, datasets, training configuration, and evaluation prompts needed to reproduce the reported experiments.

\item {\bf Experimental setting/details}
    \item[] Question: Does the paper specify all the training and test details (e.g., data splits, hyperparameters, how they were chosen, type of optimizer) necessary to understand the results?
    \item[] Answer: \answerYes{} 
    \item[] Justification: \S\ref{sec:setup} reports the SFT and RL training settings and evaluation protocol, while the appendix documents datasets and evaluation prompts.

\item {\bf Experiment statistical significance}
    \item[] Question: Does the paper report error bars suitably and correctly defined or other appropriate information about the statistical significance of the experiments?
    \item[] Answer: \answerNo{} 
    \item[] Justification: The reported tables and figures provide point estimates but do not include error bars, confidence intervals, or statistical significance tests.

\item {\bf Experiments compute resources}
    \item[] Question: For each experiment, does the paper provide sufficient information on the computer resources (type of compute workers, memory, time of execution) needed to reproduce the experiments?
    \item[] Answer: \answerYes{} 
    \item[] Justification: \S\ref{sec:setup} reports that RL training uses 32 A100 GPUs for 80 iterations, with batch size 256 and GRPO group size 16, specifying the compute hardware and scale used for the main RL experiment.
    
\item {\bf Code of ethics}
    \item[] Question: Does the research conducted in the paper conform, in every respect, with the NeurIPS Code of Ethics \url{https://neurips.cc/public/EthicsGuidelines}?
    \item[] Answer: \answerYes{} 
    \item[] Justification: The experiments use existing benchmark datasets and do not recruit human participants; the paper reports no deviation from the NeurIPS Code of Ethics.

\item {\bf Broader impacts}
    \item[] Question: Does the paper discuss both potential positive societal impacts and negative societal impacts of the work performed?
    \item[] Answer: \answerYes{} 
    \item[] Justification: The Impact Statements section discusses potential benefits for visual question answering and OCR and notes that improved visual reasoning could also be used in sensitive domains.
    
\item {\bf Safeguards}
    \item[] Question: Does the paper describe safeguards that have been put in place for responsible release of data or models that have a high risk for misuse (e.g., pre-trained language models, image generators, or scraped datasets)?
    \item[] Answer: \answerNA{} 
    \item[] Justification: The paper does not describe releasing a high-risk dataset, pretrained model, or deployed system that would require release safeguards.

\item {\bf Licenses for existing assets}
    \item[] Question: Are the creators or original owners of assets (e.g., code, data, models), used in the paper, properly credited and are the license and terms of use explicitly mentioned and properly respected?
    \item[] Answer: \answerNo{} 
    \item[] Justification: The paper cites the datasets and prior methods it uses, including in \S\ref{sec:setup} and the appendix, but does not explicitly list their licenses or terms of use.

\item {\bf New assets}
    \item[] Question: Are new assets introduced in the paper well documented and is the documentation provided alongside the assets?
    \item[] Answer: \answerNA{} 
    \item[] Justification: The manuscript does not describe releasing a new dataset, code package, or model asset; it presents a method and experimental results.

\item {\bf Crowdsourcing and research with human subjects}
    \item[] Question: For crowdsourcing experiments and research with human subjects, does the paper include the full text of instructions given to participants and screenshots, if applicable, as well as details about compensation (if any)? 
    \item[] Answer: \answerNA{} 
    \item[] Justification: The study uses existing datasets and does not conduct crowdsourcing or collect new data from human participants.

\item {\bf Institutional review board (IRB) approvals or equivalent for research with human subjects}
    \item[] Question: Does the paper describe potential risks incurred by study participants, whether such risks were disclosed to the subjects, and whether Institutional Review Board (IRB) approvals (or an equivalent approval/review based on the requirements of your country or institution) were obtained?
    \item[] Answer: \answerNA{} 
    \item[] Justification: The study does not involve human-subject experiments or recruit participants, so IRB review is not applicable.

\item {\bf Declaration of LLM usage}
    \item[] Question: Does the paper describe the usage of LLMs if it is an important, original, or non-standard component of the core methods in this research? Note that if the LLM is used only for writing, editing, or formatting purposes and does \emph{not} impact the core methodology, scientific rigor, or originality of the research, declaration is not required.
    \item[] Answer: \answerYes{} 
    \item[] Justification: \S\ref{sec:datasets} states that GPT-4 is used during linguistic annotation to generate or refine training questions, answers, and reasoning steps; the proposed method itself is described in \S\ref{sec:method}.

\end{enumerate}

\end{document}